\documentclass{article}

\usepackage{PRIMEarxiv}

\usepackage[utf8]{inputenc} 
\usepackage[T1]{fontenc}    
\usepackage{url}            
\usepackage{booktabs}       
\usepackage{amsfonts}       
\usepackage{nicefrac}       
\usepackage{microtype}      
\usepackage{lipsum}
\usepackage{fancyhdr}       

\usepackage{multirow} 
\usepackage{booktabs} 
\usepackage{wrapfig} 
\usepackage{bm} 
\usepackage{amsmath} 
\usepackage{float}
\usepackage{tabularx}
\usepackage{algorithm}
\usepackage{algpseudocode}

\usepackage{graphicx}
\usepackage[colorlinks=true,linkcolor=blue,citecolor=blue,urlcolor=blue]{hyperref}
\graphicspath{{media/}}     

\title{Predictive-Coding-Based Autonomous Regulation of Internally
Generated and Externally Coupled Processing in Human--Robot Interaction}

\author{
  Henrique Oyama\\
  Cognitive Neurorobotics Research Unit \\
  Okinawa Institute of Science and Technology Graduate University \\
  Okinawa\\
  \texttt{henrique.oyama@oist.jp} \\
  \And
    Hiroki Sawada \\
  Cognitive Neurorobotics Research Unit \\
  Okinawa Institute of Science and Technology Graduate University \\
  Okinawa\\
  \texttt{hiroki.sawada1@oist.jp} \\
   \And
  Jun Tani \\
  Cognitive Neurorobotics Research Unit \\
  Okinawa Institute of Science and Technology Graduate University \\
  Okinawa\\
  \texttt{jun.tani@oist.jp} \\
}

\begin{document}
\maketitle

\begin{abstract}
Predictive coding characterizes adaptive behavior as a dynamic balance between internally generated predictions and external sensory evidence, yet how an embodied cognitive system can regulate this balance online during ongoing interaction remains poorly understood. This study proposes a predictive-coding-based mechanism for regulating internally generated and externally coupled processing during physical human--robot interaction. The framework employs a predictive-coding-inspired variational recurrent neural network (PV-RNN), in which a meta-prior controls the degree to which posterior inference is constrained by learned prior dynamics. We extend this architecture with an online mechanism that uses reconstruction error accumulated over recent interaction history to select between predefined meta-prior regimes. The mechanism was evaluated across three physical human--robot interaction tasks involving fixed structured, changing structured, and less-constrained interaction. Across all tasks, lower meta-prior values produced the expected increase in posterior--prior divergence and reduction in reconstruction error. More importantly, reconstruction-history-driven regime selection was also associated with reduced prospective prediction error and robot-side physical interaction conflict, demonstrating consequences beyond the retrospective reconstruction objective itself. Task~3 further showed that recent sensory observations can be successfully accommodated while subsequent human motion still departs from the model's prior-generated future trajectory. Overall, these findings show that accumulated reconstruction mismatch can provide an endogenous signal for regulating how strongly subsequent inference relies on learned internal dynamics relative to ongoing sensory input during embodied interaction.
\end{abstract}



\section{Introduction}
\label{sec:1:introduction}

Adaptive behavior requires an embodied agent to balance
internally generated expectations with sensory evidence arising from
ongoing interaction. Predictive coding provides a computational account
of this process in which top-down predictions are compared with bottom-up
sensory signals and resulting prediction errors drive inference
\cite{rao1999,friston2005theory}. Within the free energy principle (FEP) and
active inference frameworks, perception and action can similarly be
understood as inference under a generative model
\cite{Fri10b,Friston2017AIF,Parr2021}. Adaptive behavior therefore
depends not only on reducing sensory mismatch, but also on regulating the
relative influence of prior expectations and sensory evidence
\cite{clark2015,Feldman2009,Sterzer2018}.

This balance is particularly important in embodied systems. Excessive
reliance on learned internal dynamics may prevent adaptation to unexpected
environmental changes, whereas excessive sensitivity to sensory
fluctuations may disrupt previously learned temporal organization.
Predictive processing accounts relate this balance to the precision or
confidence assigned to predictions and sensory evidence
\cite{Feldman2009,Adams2013,Kiverstein2019}. Related robotic frameworks
have investigated closed-loop coupling among generative models, the body,
and the environment \cite{Schillaci2020,Lanillos2018,ciria2021predictive},
while cognitive neurorobotics emphasizes the acquisition and use of
internal models through embodied sensorimotor interaction
\cite{Tani_White2022}.

Predictive-coding-inspired variational recurrent neural networks (PV-RNNs) provide a framework for investigating this relationship in temporally extended sensorimotor behavior \cite{Reza2019}. Hierarchical organization is particularly relevant in predictive coding, where interactions across representational levels can support the integration of information over different levels of abstraction and temporal structure \cite{queisser2025deep}.  PV-RNN
combines hierarchical recurrent dynamics with variational inference, such that learned prior dynamics generate temporal predictions while posterior inference accommodates sensory observations. A meta-prior weights posterior--prior divergence relative to sensory reconstruction and regulates this balance: stronger regularization constrains
inference toward learned prior dynamics, whereas weaker regularization
permits greater sensory-driven posterior adaptation \cite{frontiers_oyama2025}.

This prior--sensory balance has direct consequences for embodied and
social interaction. Predictive and free-energy-based robotic studies have
examined imitation, turn-taking, sensorimotor adaptation, and coordination
\cite{wirkuttis2023turn,Tani_White2022}. In physical human--robot
interaction, PV-RNN has been combined with compliant force-feedback
control to investigate how learned generative dynamics interact with
human-imposed motion \cite{hiroki2025}, while related active inference
approaches have examined the balance between goal-directed behavior and
adaptation to an interaction partner \cite{Murata2026}. Together, these
studies establish the importance of regulating the influence of internal
predictions and external sensory information in embodied interaction.

A key unresolved question is how an embodied predictive system can
determine online when to rely more strongly on learned internal dynamics
and when to become more strongly coupled to sensory input. Existing approaches generally examine a given precision, meta-prior, or interaction regime, or externally specify its change. However, interaction conditions can evolve continuously, motivating mechanisms that can adjust this balance online according to ongoing interaction dynamics. Persistent mismatch with reliable sensory observations may favor stronger sensory accommodation, whereas preserving prior dynamics may be advantageous when sensory information is unreliable or unavailable. These considerations motivate mechanisms for regulating this balance online during embodied interaction.

Previous PV-RNN work suggests that such regulation can arise from the
system's own inference dynamics: the meta-prior was adapted according to
reconstruction error accumulated over recent observations
\cite{frontiers_oyama2025}. Here, reconstruction error denotes the
mismatch between sensory observations and their reconstructions obtained
through posterior inference. Its temporal evolution may serve
not only as a local signal for updating inferred states, but also as a
higher-order signal regulating \emph{how} subsequent inference balances
prior expectations and sensory evidence. Whether this principle can
regulate continuous physical interaction, where inference and prediction
have immediate behavioral consequences, remains to be established.

In this study, we propose a predictive-coding-based mechanism in which reconstruction error accumulated over a recent temporal window drives online selection between predefined PV-RNN meta-prior regimes. Persistent reconstruction mismatch promotes a lower meta-prior, allowing greater posterior departure from learned prior dynamics and stronger sensory accommodation, whereas sufficiently low mismatch promotes a higher meta-prior and stronger prior constraint. The learned generative-model parameters remain fixed during interaction, allowing the mechanism to regulate the conditions of
inference rather than modify the internal model itself.

The mechanism is evaluated using a humanoid robot in three physical
human--robot interaction tasks: fixed structured interaction, changing
structured interaction that violates learned transition dynamics, and
continuous less-constrained interaction. We characterize the resulting
processing regimes using reconstruction error, posterior--prior
divergence, prospective prediction error, physical interaction conflict,
and latent dynamics. This enables us to distinguish accommodation of
recent sensory observations from prediction and coordination with the
interaction partner.

The main contributions of this work are threefold. First, we introduce an embodied predictive-coding mechanism in which the recent history of reconstruction mismatch provides an endogenous signal for online selection between predefined inference regimes with different degrees of prior constraint. Second, we show that this regulation has consequences beyond the expected reconstruction--KL trade-off, affecting prospective prediction and robot-side physical interaction. Third, we demonstrate the operation of the mechanism across fixed structured, changing structured, and less-constrained human--robot interaction, including conditions in which successful accommodation of recent sensory observations does not ensure accurate prediction of subsequent interaction.


\section{Predictive-Coding Framework for Autonomous Regulation}
\label{sec:2}

\subsection{Predictive-Coding-Inspired Variational Recurrent Neural Network}
\label{sec:2:rnn}

The proposed framework is based on a predictive-coding-inspired
variational recurrent neural network (PV-RNN) \cite{Reza2019}, which
combines hierarchical recurrent dynamics with variational inference.
Learned prior distributions generate temporal predictions, whereas
approximate posterior distributions are inferred from sensory observations
through free energy minimization.

At each time step $t$, each hierarchical layer $l$ contains deterministic
latent states $\mathbf{d}^{l}_{t}$ and stochastic latent states
$\mathbf{z}^{l}_{t}$. The prior distribution of the stochastic latent
variables is

\begin{equation}
p_{\theta}
\left(
\mathbf{z}^{l}_{t}
\mid
\mathbf{d}^{l}_{t-1}
\right)
=
\mathcal{N}
\left(
\boldsymbol{\mu}^{p,l}_{t},
\left(
\boldsymbol{\sigma}^{p,l}_{t}
\right)^2
\right)
\label{eq:prior}
\end{equation}
where $\boldsymbol{\mu}^{p,l}_{t}$ and
$\boldsymbol{\sigma}^{p,l}_{t}$ are generated from the preceding
deterministic state. During inference, the approximate posterior is

\begin{equation}
q_{\phi}
\left(
\mathbf{z}^{l}_{t}
\mid
\mathbf{d}^{l}_{t-1},
\mathbf{A}^{l}_{t}
\right)
=
\mathcal{N}
\left(
\boldsymbol{\mu}^{q,l}_{t},
\left(
\boldsymbol{\sigma}^{q,l}_{t}
\right)^2
\right)
\label{eq:posterior}
\end{equation}
where $\mathbf{A}^{l}_{t}$ denotes the adaptive variables that parameterize
the approximate posterior and are optimized through error regression using
sensory observations within the current inference window
\cite{Reza2019}. The deterministic hidden and latent states are then updated recurrently across hierarchical layers as

\begin{equation}
\begin{aligned}
\mathbf{h}^{l}_{t}
={}&
\left(1-\frac{1}{\tau^{l}}\right)\mathbf{h}^{l}_{t-1}
+
\frac{1}{\tau^{l}}
\Big(
\mathbf{W}^{ll}_{dd}\mathbf{d}^{l}_{t-1}
+
\mathbf{W}^{ll}_{zd}\mathbf{z}^{l}_{t}
\\
&+
\mathbf{W}^{l,l+1}_{dd}\mathbf{d}^{l+1}_{t-1}
+
\mathbf{W}^{l,l-1}_{dd}\mathbf{d}^{l-1}_{t-1}
+
\mathbf{b}^{l}_{h}
\Big)
\\
\mathbf{d}^{l}_{t}
={}&
\tanh\left(\mathbf{h}^{l}_{t}\right)
\end{aligned}
\label{eq:state}
\end{equation}
where $\tau^{l}$ is the layer-specific time constant. Different time
constants enable representation of temporal regularities at multiple
scales \cite{yamashita2008}. The deterministic states generate
the sensory reconstruction $\bar{\mathbf{X}}_{t}$ through the output
mapping of the generative model. The free energy objective is

\begin{equation}
\mathcal{F}
=
\sum_{t}
\left[
\left\|
\mathbf{X}_{t}
-
\bar{\mathbf{X}}_{t}
\right\|_{2}^{2}
+
\sum_{l}
w^{l}
D_{\mathrm{KL}}
\left(
q_{\phi}(\mathbf{z}^{l}_{t})
\parallel
p_{\theta}(\mathbf{z}^{l}_{t})
\right)
\right]
\label{eq:loss}
\end{equation}
where $D_{\mathrm{KL}}(q\|p)$ denotes the Kullback--Leibler (KL)
divergence from the posterior distribution $q$ to the prior distribution
$p$, and $w^l$ denotes the meta-prior for layer $l$, following notation from \cite{Reza2019}. Although the formulation permits layer-specific meta-priors, in the present study the same meta-prior is applied to all layers; hence, $w^l=w$ for all $l$. The first term quantifies sensory reconstruction mismatch, whereas the second constrains the inferred posterior distribution relative to the
learned prior. During training, the network parameters and adaptive posterior variables are optimized jointly. During interaction, the learned network
parameters remain fixed and only $\mathbf{A}$ is optimized within the
sliding inference window, allowing posterior states to accommodate ongoing
sensory observations without modifying the learned generative dynamics.

\subsection{Internally Generated and Externally Coupled Processing}
\label{sec:2:processing_regimes}

The meta-prior $w$, applied equally to all layers, determines the
relative contribution of posterior--prior divergence to the free energy
objective and therefore regulates how strongly posterior inference is
constrained by learned prior dynamics. A larger $w$ penalizes
posterior departure from the prior more strongly, producing what we refer
to as \emph{internally generated processing}. A smaller $w$ permits
greater posterior departure in response to sensory observations,
corresponding to \emph{externally coupled processing}.

These terms denote computational operating regimes rather than categorical
cognitive states. Neither regime is intrinsically preferable: stronger
prior constraint preserves learned temporal organization, whereas weaker
constraint enables greater accommodation of ongoing sensory information.
Previous PV-RNN studies have demonstrated substantial changes in
prior-driven and sensory-driven dynamics through meta-prior manipulation
\cite{Reza2019,hiroki2025,wirkuttis2023turn}. Here, rather than fixing the
meta-prior to a single interaction value, we regulate it online from the
recent history of reconstruction mismatch.

\subsection{Reconstruction-Error-Driven Meta-Prior Regulation}
\label{sec:2:w-adaptation}

During interaction, error regression continuously updates posterior latent
states within the inference window. Persistent mismatch over this recent
history is quantified by the temporally averaged reconstruction error:

\begin{equation}
\bar{E}_{r,t}
=
\frac{1}{T_W}
\sum_{\tau=t-T_W+1}^{t}
\left\|
\mathbf{X}_{\tau}
-
\bar{\mathbf{X}}_{\tau}
\right\|_{2}^{2}
\label{eq:avg_recc_err}
\end{equation}

This quantity measures how well posterior inference reconstructs recently
observed sensory data. Importantly, $\bar{E}_{r,t}$ is retrospective and
does not directly measure prediction accuracy for future, previously
unseen observations. This distinction is used later to compare sensory
accommodation with prospective coordination.

The accumulated reconstruction error regulates transitions between a high
meta-prior $w_H$, corresponding to stronger prior constraint, and a low
meta-prior $w_L$, corresponding to stronger sensory accommodation. With
lower and upper thresholds $\mathrm{Thr}_L$ and $\mathrm{Thr}_H$, the
probability of transitioning from $w_L$ to $w_H$ is

\begin{equation}
P
\left(
w_{t+1}=w_H
\mid
w_t=w_L
\right)
=
\sigma
\left(
\frac{
\mathrm{Thr}_{L}
-
\bar{E}_{r,t}
}{
\mathrm{Temp}
}
\right)
\label{eq:adaptW_F}
\end{equation}
whereas the probability of transitioning from $w_H$ to $w_L$ is

\begin{equation}
P
\left(
w_{t+1}=w_L
\mid
w_t=w_H
\right)
=
\sigma
\left(
\frac{
\bar{E}_{r,t}
-
\mathrm{Thr}_{H}
}{
\mathrm{Temp}
}
\right)
\label{eq:adaptW_MW}
\end{equation}
where $\sigma(a)=1/[1+\exp(-a)]$ is the logistic sigmoid function and
$\mathrm{Temp}$ controls transition sharpness. Separate lower and upper
thresholds introduce hysteresis: increasing mismatch promotes transition
toward $w_L$, whereas sufficiently low mismatch promotes return to
$w_H$. The temperature parameter makes these transitions probabilistic
rather than strictly deterministic.

Here, ``autonomous'' refers specifically to online selection between the
predefined meta-prior regimes based on the system's endogenous
reconstruction dynamics, without an external command specifying when to
switch. The regulatory policy and its hyperparameters are predefined
rather than learned autonomously.

The selected meta-prior regulates subsequent inference.
Specifically, $\bar{E}_{r,t}$ is evaluated after posterior optimization
over the current inference window; the resulting $w$ therefore changes
the conditions of the following inference step rather than retroactively
altering the posterior from which the current reconstruction error was
obtained.

\subsection{Closed-Loop Online Regulation During Interaction}
\label{sec:2:closed_loop}

The complete mechanism forms a closed regulatory loop. At each interaction
step, newly observed sensorimotor data enter the sliding inference window,
error regression updates the adaptive variables and posterior latent
states, and $\bar{E}_{r,t}$ is evaluated. This accumulated mismatch
determines the probability of changing the meta-prior regime, which then
regulates the balance between prior constraint and sensory accommodation
during subsequent inference. Algorithm~\ref{alg:adaptive_switching}
summarizes the procedure.

\begin{algorithm}[!t]
\caption{Online Reconstruction-Error-Driven Meta-Prior Regulation}
\label{alg:adaptive_switching}
\begin{algorithmic}[1]
\State Initialize PV-RNN states and meta-prior $w$
\For{each interaction step $t$}
    \State Acquire sensory observation $\mathbf{X}_{t}$
    \State Update the sliding inference window
    \State Optimize adaptive variables $\mathbf{A}$ by error regression
    \State Infer posterior latent states and reconstruct observations
    \State Compute $\bar{E}_{r,t}$ using \eqref{eq:avg_recc_err}
    \If{$w=w_L$}
        \State Compute $P(w_{t+1}=w_H)$ using \eqref{eq:adaptW_F}
        \State Sample the transition from $w_L$ to $w_H$
    \ElsIf{$w=w_H$}
        \State Compute $P(w_{t+1}=w_L)$ using \eqref{eq:adaptW_MW}
        \State Sample the transition from $w_H$ to $w_L$
    \EndIf
    \State Apply the resulting $w$ to subsequent inference
\EndFor
\end{algorithmic}
\end{algorithm}

\begin{figure}[t]
    \centering
    \includegraphics[width=0.8\columnwidth]{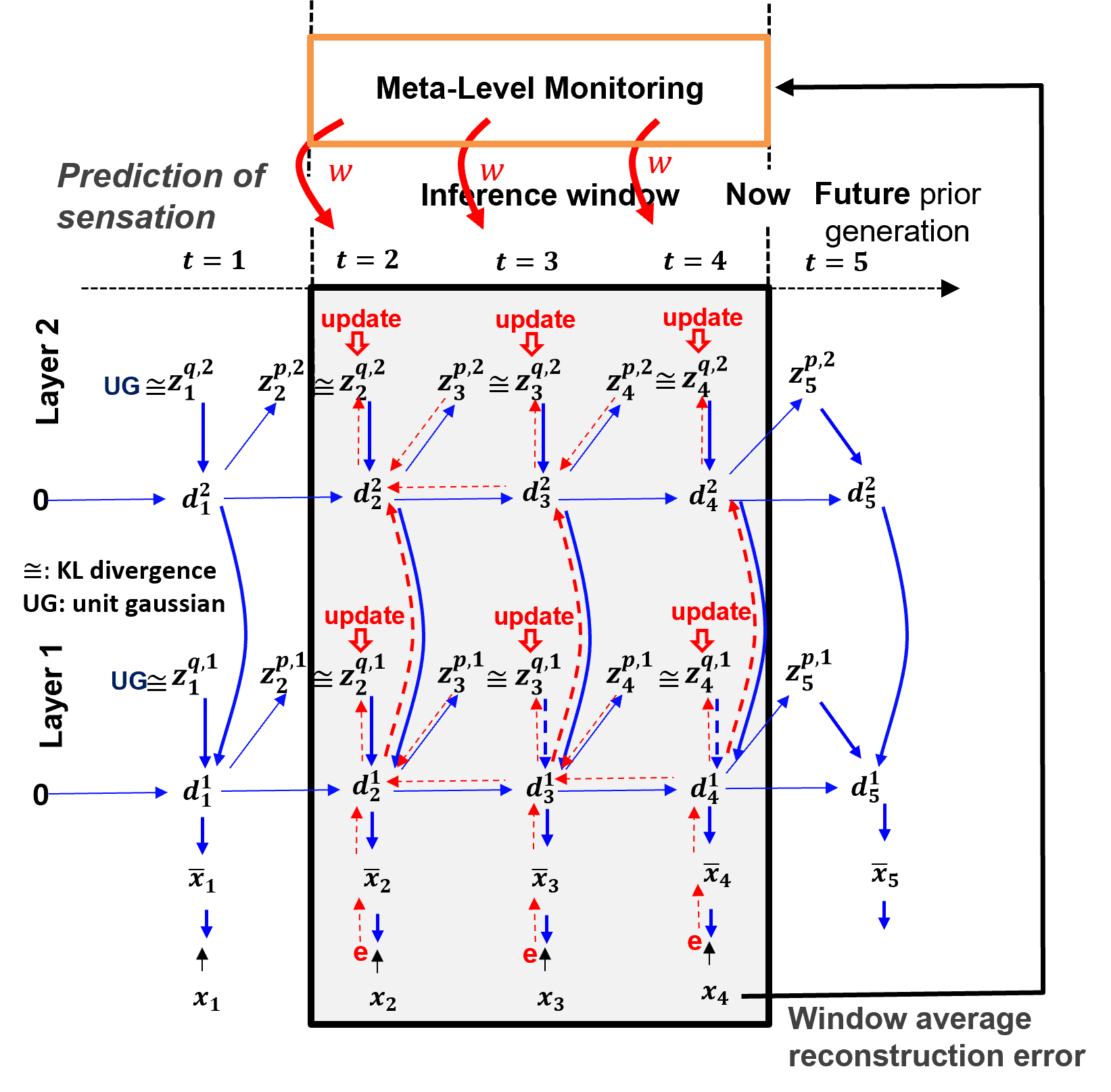}
    \caption{Hierarchical two-layer PV-RNN architecture with online
    meta-prior modulation based on temporally accumulated reconstruction
    error. Solid blue lines represent the generative process, while dotted
    red lines indicate posterior inference. The shaded region schematically
    illustrates the sliding inference window used to compute reconstruction
    error $\bar{E}_{r,t}$; the experiments use $T_W=100$. The accumulated
    reconstruction error regulates probabilistic transitions between
    internally generated and externally coupled processing.}
    \label{fig:PV-RNN}
\end{figure}

The hierarchical implementation is illustrated in
Fig.~\ref{fig:PV-RNN}. The shaded region denotes the sliding inference
window in which posterior states are updated from sensory observations.
The reconstruction error averaged over this window is passed to the
meta-level regulatory mechanism, which modulates the meta-prior for
subsequent inference, while the generative process extends beyond the
current window to produce future prior dynamics.

Thus, reconstruction mismatch has two related computational roles: it
drives posterior optimization within the current inference window, while
its temporal accumulation regulates how strongly subsequent inference is
constrained by the learned prior. This higher-order regulatory role enables
online adaptation of the prior--sensory balance without changing the
learned network parameters.


\section{Experimental Methods}
\label{sec:3}

\subsection{Humanoid Robot and Sensorimotor Variables}
\label{sec:3:robot}

The proposed framework was evaluated using the Torobo humanoid robot in
physical human--robot interaction. Torobo has 16 degrees of freedom (DoF),
including 6 DoF in each arm, and supports compliant kinesthetic
interaction through force-feedback control. Four arm-joint angles, two
from each arm, constituted the sensorimotor observations
$\mathbf{X}_t$ processed by the PV-RNN.

\begin{figure*}[!t]
    \centering
    \includegraphics[width=\columnwidth]{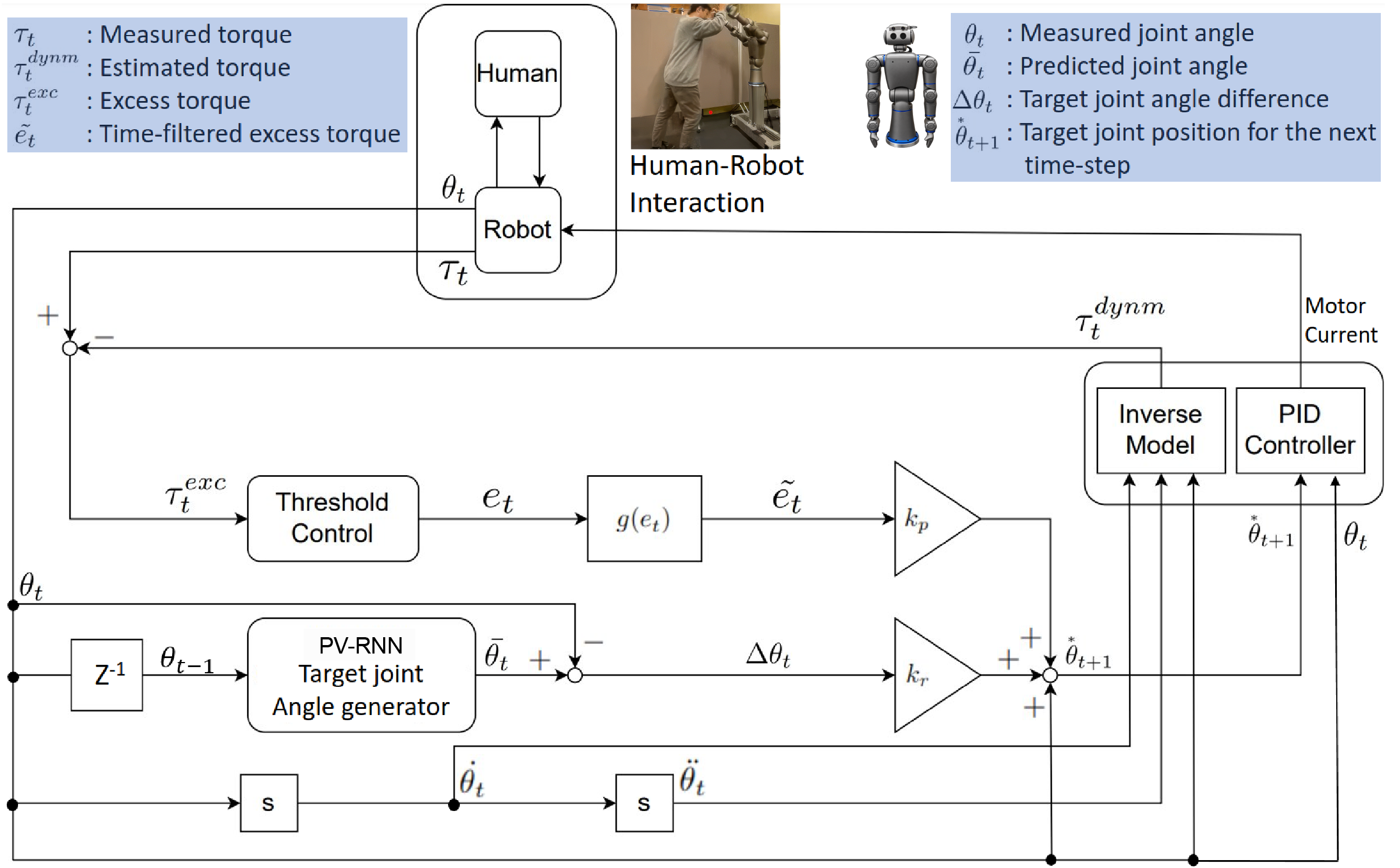}
    \caption{Control architecture for physical human--robot interaction     (adapted from \cite{hiroki2025}). The PV-RNN generates target joint trajectories while force-feedback control accommodates externally imposed
    human motion, enabling continuous closed-loop interaction between
    generative dynamics and sensory input. See \cite{hiroki2025} for a
    detailed description of the control architecture and notation.}
    \label{fig:controller}
\end{figure*}

The control architecture is shown in Fig.~\ref{fig:controller}. The PV-RNN generates target joint trajectories from its inferred sensorimotor dynamics, which are integrated with the robot's force-feedback controller. An author, who provided informed consent for participation and publication of the resulting data, acted as the human operator and physically guided the robot while the PV-RNN continuously performed online inference, forming a closed loop between learned generative dynamics and sensory input. 

Physical interaction conflict was quantified from the robot's recorded excess-torque signal, expressed in N$\cdot$m. Larger values indicate stronger disagreement between the robot's internally generated motor command and externally imposed motion. This controller-level quantity is derived from the robot's internal control signals and was used only as a robot-side measure of relative physical conflict between processing regimes; it is not a measurement of human force or an independently calibrated biomechanical interaction-force measure.

\subsection{Training Data and Model Training}
\label{sec:3:training}

The PV-RNN was trained on four-dimensional joint-angle trajectories
representing sensorimotor dynamics subsequently encountered during
physical interaction. Tasks~1 and 2 used probabilistically structured
sequences generated from learned movement primitives, whereas Task~3 used
less-constrained trajectories collected during continuous kinesthetic
human--robot interaction.

\subsubsection{Structured Training Data for Tasks 1 and 2}
\label{sec:3:training_structured}

\begin{figure}[t]
    \centering 
    \includegraphics[width=0.75\columnwidth]{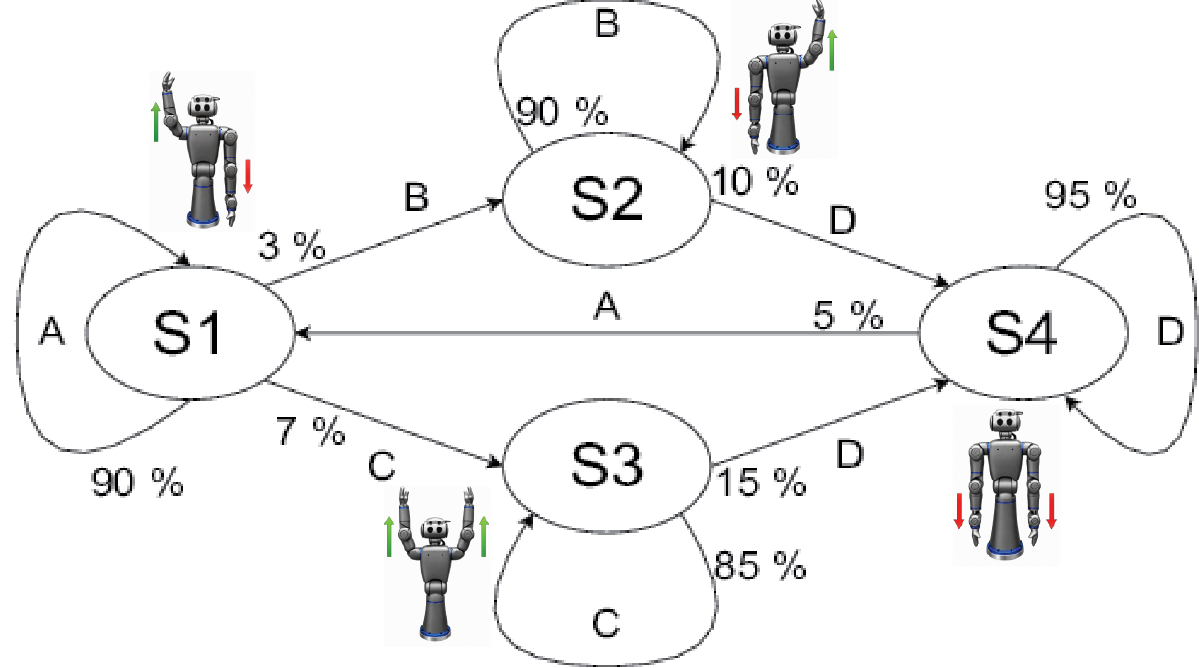}
    \caption{Probabilistic finite state machine used to generate the
    structured training sequences for Tasks~1 and 2 (adapted from \cite{hiroki2025}). The four states correspond to movement primitives A--D, with transitions governed by the indicated probabilities.}
    \label{fig:PFSM}
\end{figure}

For Tasks~1 and 2, training sequences were organized according to the
probabilistic finite state machine (PFSM) shown in Fig.~\ref{fig:PFSM},
following the interaction framework of \cite{hiroki2025}. Four cyclic
movement primitives, denoted A--D, were defined with a cycle length of
20 time steps. The PFSM specifies stochastic transitions among these
primitives at the end of each cycle, combining stable local periodic
dynamics with probabilistic transitions over longer temporal scales.

Ten training sequences were generated, each containing 200 movement
cycles (4000 time steps). Four PV-RNN models were independently trained
using identical hyperparameters but different random seeds to evaluate
robustness across model initialization.

\subsubsection{Less-Constrained Training Data for Task 3}
\label{sec:3:training_task3}

For Task~3, training data were obtained through direct kinesthetic
interaction between a human operator and the humanoid robot without a
predefined PFSM. Four datasets of approximately 1300--1500 time steps
were recorded during continuous interaction. The same author-operator guided the robot arms using movement patterns broadly similar to those represented in the structured data while allowing timing, amplitude, and transitions to vary
naturally. Consequently, these trajectories exhibited weaker temporal
regularity and greater variability than the PFSM-generated sequences,
allowing the PV-RNN to learn broader sensorimotor regularities without an
explicit transition schedule.

\subsubsection{Network Architecture and Training}
\label{sec:3:network_training}

The PV-RNN contained two hierarchical recurrent layers. As summarized in
Table~\ref{tab:training_params}, the upper layer (or layer 2) contained 30
deterministic units and four stochastic latent variables with
$\tau^{2}=6$, whereas the lower layer (or layer 1) contained 60 deterministic units and two stochastic latent variables with $\tau^{1}=3$. The training meta-prior was fixed at $w^{\mathrm{tr}}=0.01$ and applied equally to both PV-RNN layers throughout training.

\begin{table}[t]
\centering
\caption{PV-RNN training parameters.}
\label{tab:training_params}
\begin{tabular}{lcccc}
\hline
Layer & $\#d$ & $\#z$ & $\tau^{l}$ & $w^{\mathrm{tr}}$ \\
\hline
Layer 2 (Top)    & 30 & 4 & 6 & 0.01 \\
Layer 1 (Bottom) & 60 & 2 & 3 & 0.01 \\
\hline
\end{tabular}
\end{table}

Training was performed for 50,000 epochs using the Adam optimizer
\cite{kingma2014adam} with a learning rate of 0.001. Model parameters and
adaptive posterior variables were optimized using the variational
objective in Section~\ref{sec:2:rnn}. The meta-prior remained fixed
during training and was regulated only during the subsequent interaction
experiments; thus, the online regulatory switching policy was not
explicitly learned from the training data.

\subsection{Human--Robot Interaction Tasks}
\label{sec:3:tasks}

Three physical human--robot interaction tasks varied the temporal
relationship between the robot's learned dynamics and externally imposed
human motion. In all tasks, online posterior inference used a sliding
error-regression window of $T_W=100$ time steps, with the meta-prior
regulated by the mechanism in Section~\ref{sec:2:w-adaptation}. Tasks~1
and 2 used movement primitives learned from the PFSM, whereas Task~3
examined continuous less-constrained interaction.

During online interaction, the same meta-prior value was applied to both
PV-RNN layers. Tasks~1 and 2 used $w_H=1.0$, $w_L=0.01$, $Thr_L=0.0005$, $Thr_H=0.040$, and $Temp=0.005$, whereas Task~3 used $w_H=1.0$, $w_L=0.001$, $Thr_L=0.005$, $Thr_H=0.040$, and $Temp=0.001$. These regulatory parameters were calibrated after preliminary inspection of interaction runs to obtain temporally distinguishable high- and low-$w$ regimes. Following calibration, the Tasks~1/2 parameter set was held fixed across all four independently trained models, as was the Task~3 parameter set. Thus, parameters were task-specific but were not adjusted separately for individual model realizations during evaluation. Once calibrated, these parameters define a fixed switching policy for each task condition, whereas the timing of individual regime transitions is determined online from the evolving reconstruction-error history.

\subsubsection{Task 1: Fixed Structured Interaction}
\label{sec:3:task1}

Each Task~1 block consisted of 100 time steps of free generation followed
by 200 time steps of physical interaction. During interaction, the human
guided the robot according to one learned primitive (A--D), which remained
fixed throughout the 200-step period. This 300-step sequence was repeated
ten times, yielding 3000 time steps per experiment. Task~1 therefore
produced sustained structured mismatch whenever the human-imposed
trajectory differed from the robot's internally generated dynamics.

\subsubsection{Task 2: Changing Structured Interaction}
\label{sec:3:task2}

Task~2 retained the same 100-step free-generation and 200-step interaction
structure, but the human-imposed primitive changed every 100 interaction
steps. These transitions were selected not to follow the probabilistic
transition structure encountered during training, repeatedly violating the
robot's learned temporal expectations. Task~2 therefore produced recurring
structured mismatch rather than the comparatively stationary mismatch of
Task~1. 

\subsubsection{Task 3: Less-Constrained Human--Robot Interaction}
\label{sec:3:task3}

Task~3 consisted of continuous less-constrained physical interaction
without predefined free-generation segments or a prescribed schedule of
movement-primitive transitions. The same author-operator continuously guided the robot arms using movement patterns broadly compatible with the Task~3 training data, while timing, amplitude, and transitions varied naturally. Each
interaction sequence lasted approximately 900--1200 time steps.

The term \emph{less-constrained} is used because the interaction remained
limited by the robot morphology, the four sensorimotor dimensions, and
the physical control architecture; only the explicit temporal prescription
of movement primitives and transition timing used in Tasks~1 and 2 was
removed. 

\subsection{Evaluation Metrics and Statistical Analysis}
\label{sec:3:evaluation}

Interaction dynamics were evaluated using complementary measures of inference, prospective prediction, robot-side physical interaction, and latent-state organization. Samples were classified according to whether the active meta-prior corresponded to the high-$w$ or low-$w$ regime.

Reconstruction error quantified accommodation of observations already
available within the error-regression window, using both instantaneous
error and the temporally averaged $\bar{E}_{r,t}$ in (\ref{eq:avg_recc_err}). The KL-divergence terms of the upper and lower PV-RNN layers were evaluated to quantify departure of the inferred posterior from the learned prior. We report the raw KL divergence, which directly measures posterior--prior departure independently of the meta-prior weighting applied to the KL term in the optimization objective.

Prospective prediction error was evaluated separately from reconstruction error. At each online step, the first prior-generated sensory sample beyond the error-regression window was compared with the newest sensory observation at the following step, without access to that observation during prediction. Error was quantified as the mean absolute difference across sensorimotor dimensions. Thus, reconstruction error measures retrospective accommodation of observations available during inference, whereas prospective prediction error measures one-step-ahead anticipation of subsequently observed sensory input.

Latent state organization was analyzed using kernel principal component
analysis (kernel PCA) with a radial-basis-function kernel applied
separately to the deterministic latent states and PV-RNN output
trajectories. KPC1 and KPC2 denote the first and second kernel principal
component coordinates, respectively. The deterministic latent dynamics
were visualized in the KPC1--KPC2 plane. For the generated output
dynamics, we used a phase-space representation of KPC1 versus its
one-step temporal change, $\Delta\mathrm{KPC1}_t=\mathrm{KPC1}_{t+1}-\mathrm{KPC1}_t$, to visualize both the projected output state and its local temporal evolution.

For direct comparison between Tasks~1 and 2, a common kernel-PCA
embedding was fitted to pooled samples from both tasks before projecting
the task-specific trajectories. Task~3 was embedded separately because
it was trained from a different, less-constrained sensorimotor dataset.
The percentages shown with the kernel principal components indicate
their relative eigenvalue contributions in the centered kernel matrix.

Statistical comparisons between high- and low-$w$ processing were performed across the four independently trained model realizations. Individual time steps were not treated as independent replicates; each metric was first aggregated within each model and meta-prior regime, and the resulting model-level values constituted paired observations. Across-model means and 95\% confidence intervals were calculated using the Student-$t$ distribution. Paired high-$w$--low-$w$ effects were summarized by the mean within-model difference and its 95\% confidence interval. Given the small number of model realizations ($n=4$), these intervals are reported as descriptive estimates of across-model variability rather than as evidence from formal hypothesis testing. For Tasks~1 and 2, statistical summaries were restricted to physical-interaction periods, excluding free-generation intervals and the first five time steps following each meta-prior transition.


\section{Results}
\label{sec:results}

We examine how reconstruction-error-driven meta-prior regulation unfolds
across the three interaction conditions. Results are reported across four
independently trained model realizations, with emphasis on paired effect
estimates and 95\% confidence intervals. Consistent with the computational
role of the meta-prior described in Section~\ref{sec:2:processing_regimes},
raw posterior--prior KL divergence is used to quantify the extent to which
posterior inference departs from the learned prior under high- and low-$w$
processing.

\subsection{Task 1: Fixed Structured Interaction}
\label{sec:4:task1}

\begin{figure*}[!t]
    \centering
    \includegraphics[width=\columnwidth]{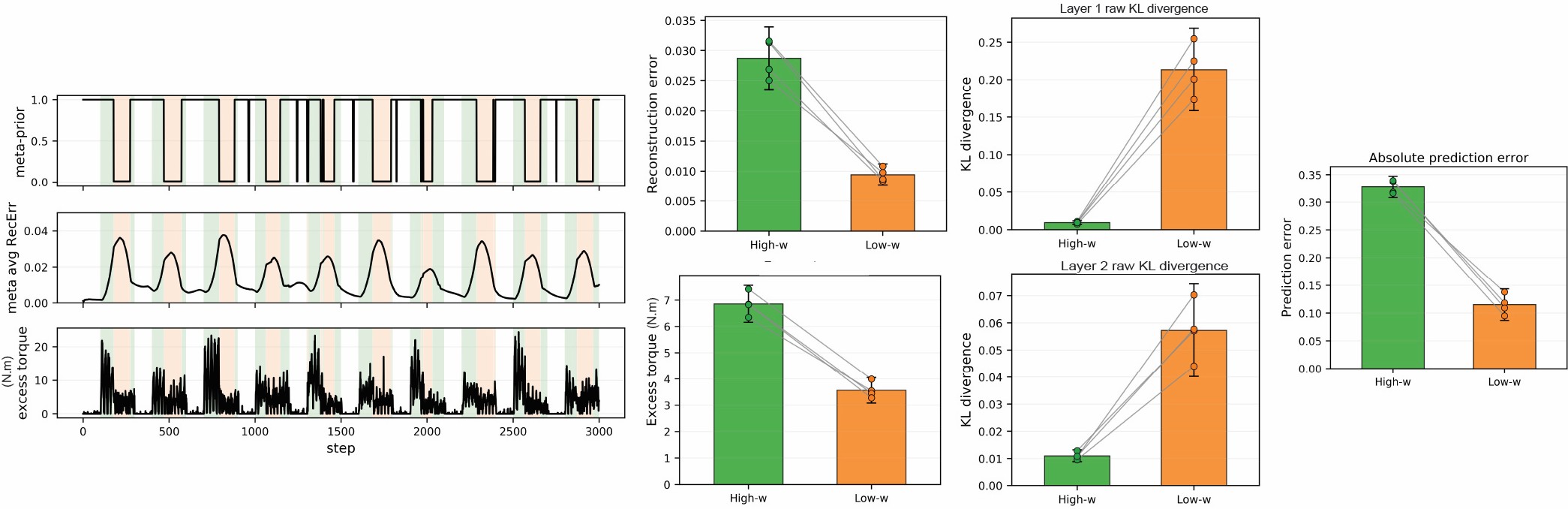}
    \caption{Online meta-prior regulation and across-model comparison
    during Task~1. Left: temporal dynamics for one representative model
    realization, showing the meta-prior, temporally averaged reconstruction
    error, and excess torque; shaded regions indicate high-$w$ (green) and
    low-$w$ (orange) processing. Right: across-model comparisons of
    reconstruction error, excess torque, raw KL divergence in Layers~1 and~2,
    and prospective prediction error under high and low $w$. Bars indicate
    across-model means, error bars indicate 95\% confidence intervals, dots
    represent independently trained models, and connecting lines indicate
    paired observations from the same model.}
    \label{fig:task1}
\end{figure*}

Figure~\ref{fig:task1} summarizes the online meta-prior regulation dynamics and across-model comparisons for Task~1. During interaction with a fixed learned movement primitive, reconstruction mismatch increased after the
onset of externally imposed interaction, promoting transitions from high
to low $w$. Under low $w$, stronger sensory accommodation subsequently
reduced reconstruction error, allowing the system to return toward
high-$w$ processing as the accumulated mismatch decreased.

Across the four models (Fig.~\ref{fig:task1}), mean reconstruction
error decreased from 0.0287 under high $w$ to 0.0094 under low $w$
(paired difference: 0.0193, 95\% CI: 0.0142--0.0244), while mean
absolute prospective prediction error decreased from 0.3276 to 0.1152
(difference: 0.2124, 95\% CI: 0.1774--0.2474). Mean raw KL divergence
increased under low $w$ in both Layer~1 (0.0094 to 0.2135) and
Layer~2 (0.0110 to 0.0573), indicating greater posterior departure
from the learned prior. Mean excess torque decreased from
6.86~N$\cdot$m to 3.57~N$\cdot$m (reduction: 3.29~N$\cdot$m,
95\% CI: 2.75--3.83~N$\cdot$m). Thus, stronger sensory accommodation
reduced reconstruction and prospective prediction errors as well as
robot-side physical interaction conflict.

\subsection{Task 2: Changing Structured Interaction}
\label{sec:4:task2}

\begin{figure*}[!t]
    \centering
    \includegraphics[width=\columnwidth]{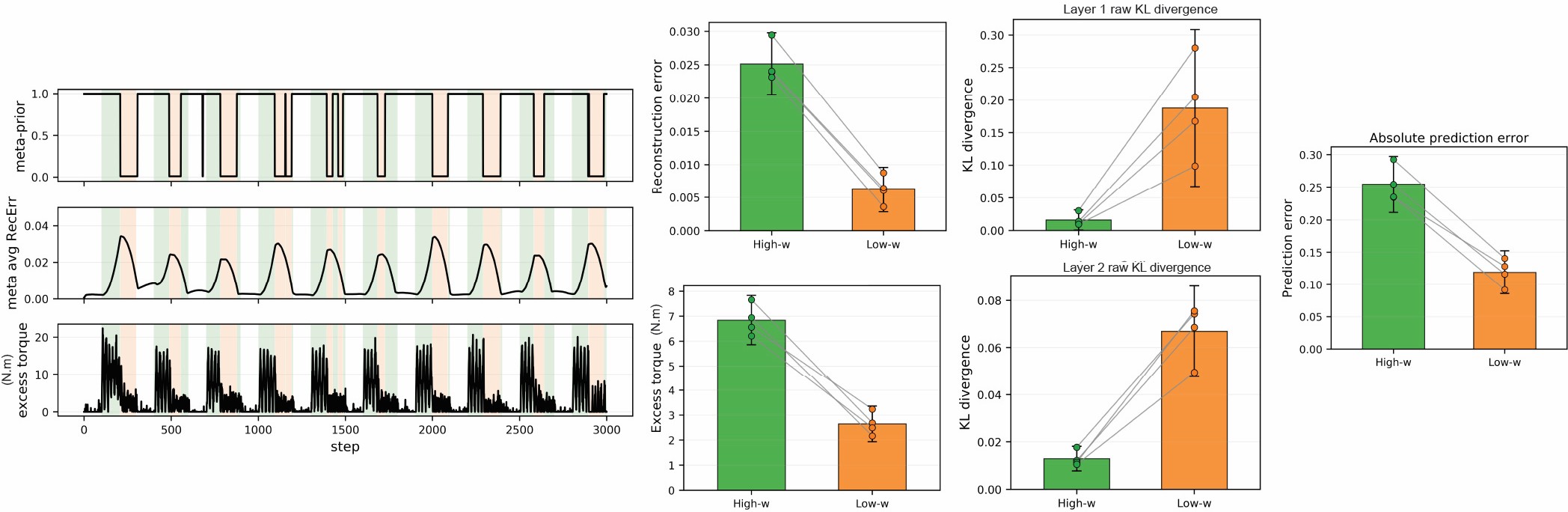}
    \caption{Online meta-prior regulation and across-model comparison
    during Task~2. Left: temporal dynamics for one representative model
    realization, showing the meta-prior, temporally averaged reconstruction
    error, and excess torque; shaded regions indicate high-$w$ (green) and
    low-$w$ (orange) processing. Right: across-model comparisons of
    reconstruction error, excess torque, raw KL divergence in Layers~1 and~2,
    and prospective prediction error under high and low $w$. Bars indicate
    across-model means, error bars indicate 95\% confidence intervals, dots
    represent independently trained models, and connecting lines indicate
    paired observations from the same model.}
    \label{fig:task2}
\end{figure*}

Figure~\ref{fig:task2} summarizes the online meta-prior regulation dynamics and across-model comparisons for Task~2. Repeated externally imposed changes in movement primitive produced recurrent reconstruction mismatch and corresponding regulation between high- and low-$w$ processing. The precise timing of these transitions varied across independently trained PV-RNN models.

Across the four models (Fig.~\ref{fig:task2}), mean reconstruction error decreased from 0.0251 under high $w$ to 0.0062 under low $w$ (paired difference: 0.0189, 95\% CI: 0.0166--0.0212), while mean absolute prospective prediction error decreased from 0.2544 to 0.1192 (difference: 0.1352, 95\% CI: 0.1038--0.1665). Mean raw KL divergence
increased under low $w$ in both Layer~1 (0.0160 to 0.1878) and Layer~2 (0.0129 to 0.0669), again indicating greater posterior departure from the learned prior. Mean excess torque decreased from 6.84~N$\cdot$m to 2.65~N$\cdot$m (reduction: 4.20~N$\cdot$m,
95\% CI: 2.92--5.47~N$\cdot$m). Thus, the same regime-level distinction was maintained despite repeated violations of the learned transition structure.

\subsection{Task 3: Less-Constrained Human--Robot Interaction}
\label{sec:4:task3}

\begin{figure*}[!t]
    \centering
    \includegraphics[width=\columnwidth]{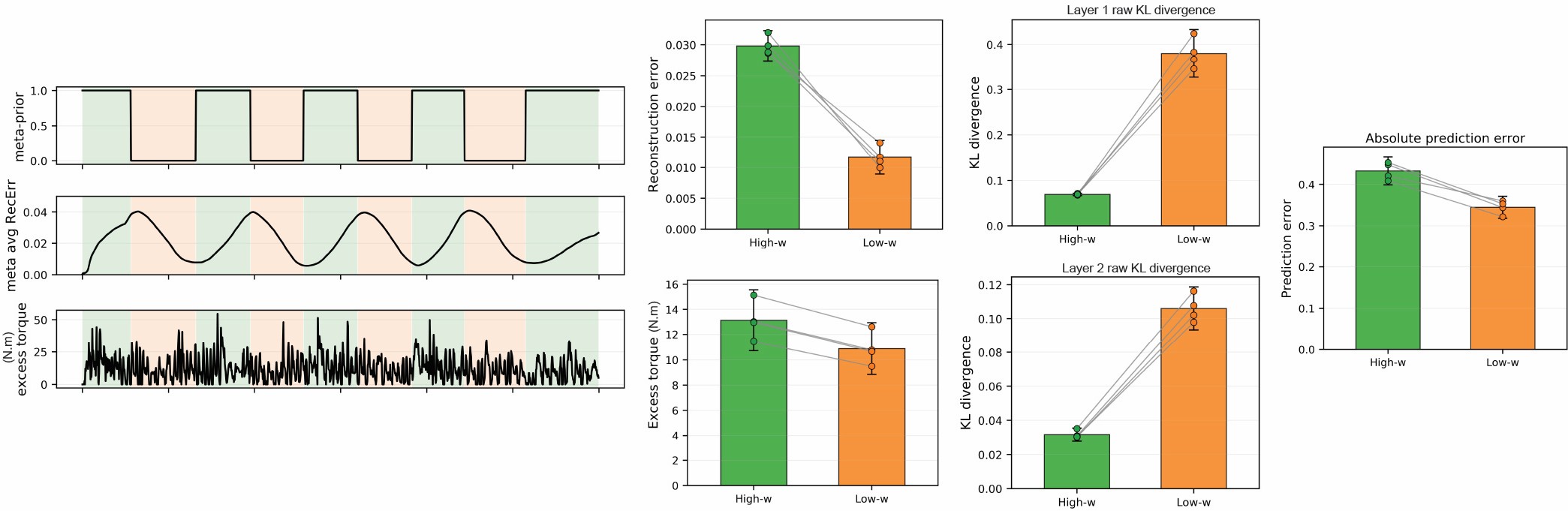}
    \caption{Online meta-prior regulation and across-model comparison
    during Task~3. Left: temporal dynamics for one representative model
    realization, showing the meta-prior, temporally averaged reconstruction
    error, and excess torque; shaded regions indicate high-$w$ (green) and
    low-$w$ (orange) processing. Right: across-model comparisons of
    reconstruction error, excess torque, raw KL divergence in Layers~1 and~2,
    and prospective prediction error under high and low $w$. Bars indicate
    across-model means, error bars indicate 95\% confidence intervals, dots
    represent independently trained models, and connecting lines indicate
    paired observations from the same model.}
    \label{fig:task3}
\end{figure*}

Figure~\ref{fig:task3} summarizes the online meta-prior regulation dynamics and across-model comparisons during less-constrained interaction. In contrast to Tasks~1 and 2, Task~3 contained no prescribed movement sequence or transition schedule. Nevertheless, reconstruction mismatch continued to drive autonomous transitions between high- and low-$w$ processing, with repeated increases in accumulated reconstruction error promoting
low-$w$ sensory accommodation.

Across the four models (Fig.~\ref{fig:task3}), mean reconstruction error decreased from 0.0298 under high $w$ to 0.0117 under low $w$ (paired difference: 0.0181, 95\% CI: 0.0133--0.0230). Mean raw KL divergence increased under low $w$ in both Layer~1
(0.0693 to 0.3800) and Layer~2 (0.0317 to 0.1060), showing that weaker meta-prior constraint again enabled greater posterior departure from the learned prior and lower reconstruction mismatch.

\subsection{Prospective Prediction and Physical Interaction in Task 3}
\label{sec:4:task3_prediction}

\begin{figure*}[!t]
    \centering
    \includegraphics[width=\columnwidth]
    {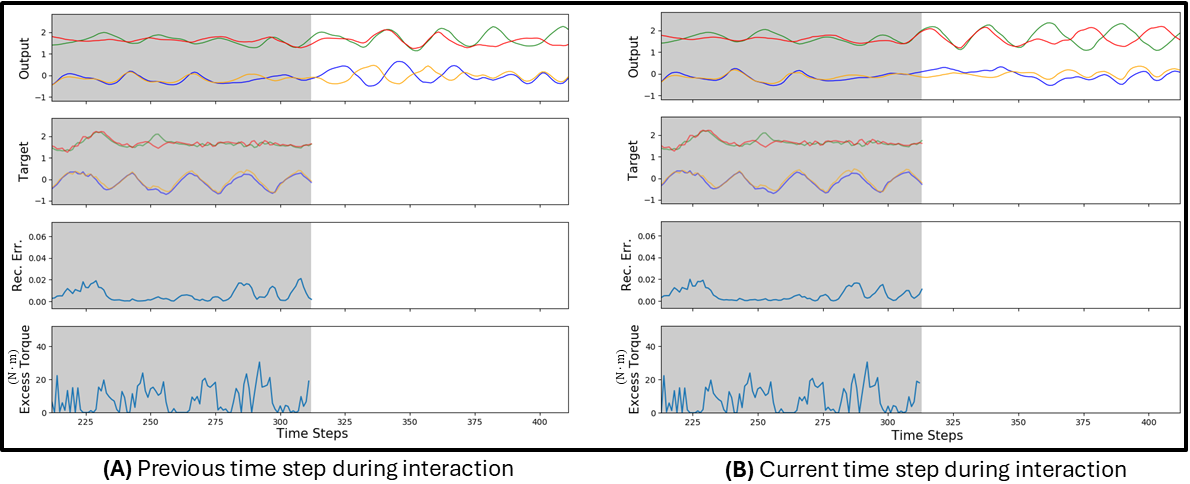}
    \caption{Relationship between prospective prediction and physical
    interaction during Task~3. Less-constrained human motion can depart
    from the robot's prior-generated future trajectory even when recent
    sensory observations are successfully accommodated through posterior
    inference.}
    \label{fig:task3_prediction}
\end{figure*}

Task~3 further distinguishes retrospective sensory accommodation from prospective coordination. As illustrated in Fig.~\ref{fig:task3_prediction}, recent human motion within the inference window can be successfully accommodated through posterior inference, while the subsequent human trajectory nevertheless departs from the
future trajectory generated from the inferred state. Thus, low reconstruction mismatch does not necessarily imply accurate anticipation of subsequent interaction dynamics.

Nevertheless, low $w$ improved prospective prediction across models (Fig.~\ref{fig:task3}): mean absolute prospective prediction error decreased from $0.4322$ to $0.3447$ (difference: $0.0875$, 95\% CI: $0.0565$--$0.1186$). Mean excess torque also decreased from $13.14$~N$\cdot$m to $10.89$~N$\cdot$m (reduction: $2.25$~N$\cdot$m, 95\% CI: $1.88$--$2.62$~N$\cdot$m). These results show that the regime-dependent effects extended beyond retrospective reconstruction to prospective prediction and robot-side physical interaction during less-constrained interaction.

\subsection{Latent-Dynamical Organization Across Interaction Conditions}
\label{sec:4:latent_dynamics}

\begin{figure*}[!t]
    \centering
    \includegraphics[width=\columnwidth]{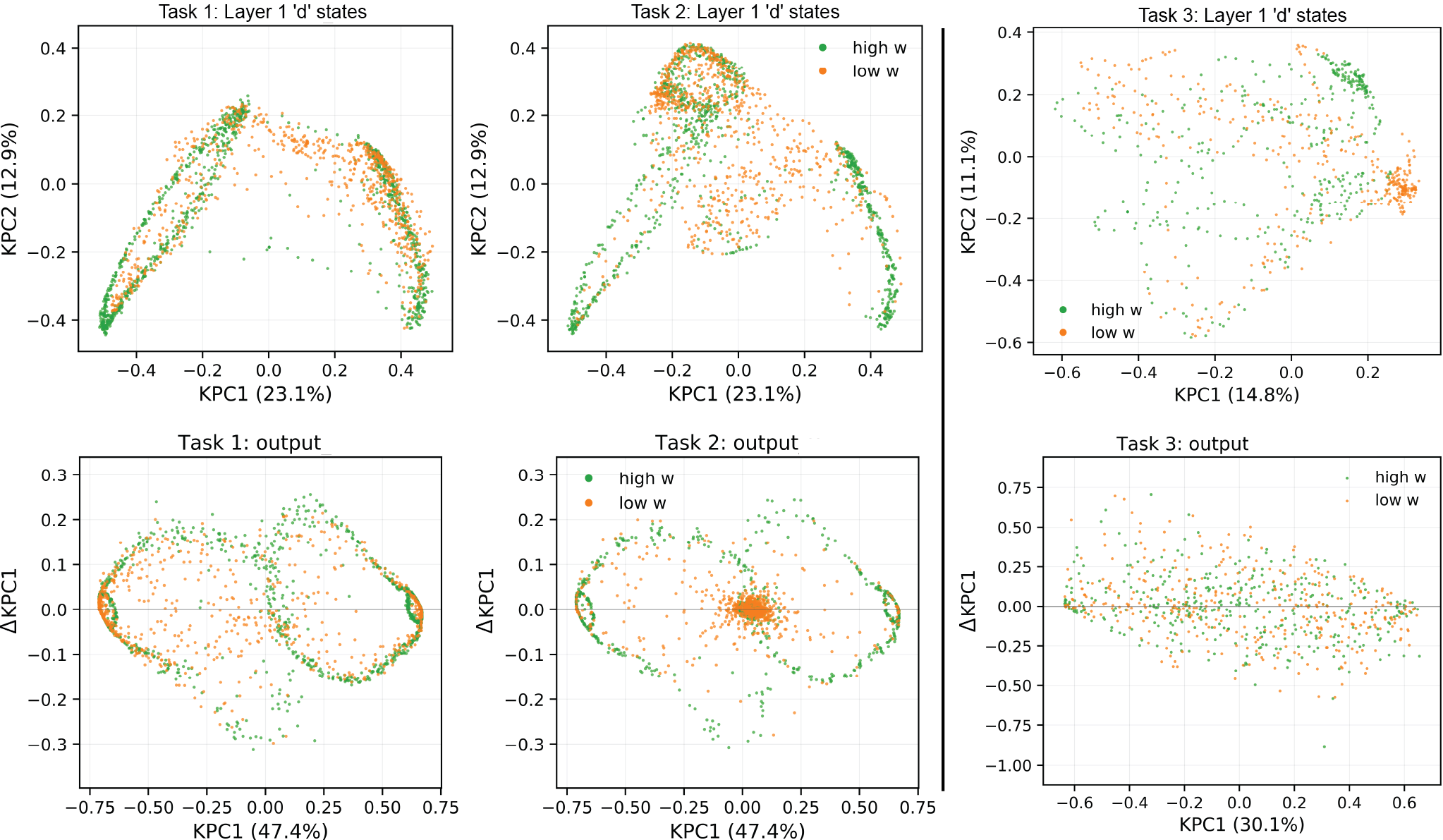}
    \caption{Kernel PCA representations of latent and output dynamics across
    the three interaction tasks for representative PV-RNN models. Top:
    deterministic latent dynamics plotted as KPC2 versus KPC1. Bottom:
    generated output dynamics plotted as the one-step change
    $\Delta\mathrm{KPC1}$ versus KPC1. Tasks~1 and 2 share a common
    kernel-PCA embedding, permitting direct comparison of their structured
    dynamical organization, whereas Task~3 was embedded separately and
    exhibits a broader, less discretely organized interaction-driven
    structure. Green and orange samples indicate high- and low-$w$
    processing, respectively. The remaining independently trained models
    exhibited qualitatively similar organization.}
    \label{fig:kpca_all_tasks}
\end{figure*}

Figure~\ref{fig:kpca_all_tasks} compares the deterministic latent and
output dynamics across the three interaction conditions for
representative PV-RNN models. Deterministic latent dynamics are shown in
the KPC1--KPC2 plane, whereas output dynamics are represented by KPC1
and its one-step temporal change $\Delta\mathrm{KPC1}$. Tasks~1 and 2
were projected using a shared kernel-PCA embedding, permitting direct
comparison of their structured dynamical organization, while Task~3 was
embedded separately.

High- and low-$w$ samples show partially different occupancy patterns
within these representations, consistent with changes in posterior
dynamics associated with meta-prior regulation.

Task~3 exhibits a qualitatively different organization. Its separately
computed kernel-PCA representation is broader and more continuously
distributed than the structured representations observed in Tasks~1 and
2, consistent with the absence of a predefined sequence of discrete
movement primitives. High- and low-$w$ samples remain partially distinguishable, particularly in the deterministic latent states, but exhibit greater overlap. Thus, meta-prior regulation continues to modulate posterior
departure from the prior within a less discretely organized,
interaction-driven state space. The remaining independently trained
models exhibited qualitatively similar organization.

\subsection{Summary Across Interaction Conditions}
\label{sec:4:summary}

Across all three tasks, low-$w$ processing produced a consistent regime-dependent signature: lower reconstruction and prospective prediction errors, greater raw posterior--prior KL divergence, and lower robot-side physical interaction conflict than high-$w$ processing. The increased posterior--prior divergence and reduced reconstruction error are consistent with the expected effect of weakening the KL constraint, whereas the accompanying differences in prospective prediction and physical interaction extend beyond this retrospective reconstruction trade-off. Because Task~3 used different regulatory parameters from Tasks~1 and~2, quantitative effect magnitudes are interpreted within each task rather than compared directly across tasks.


\section{Discussion}
\label{sec:discussion}

This study investigated whether the recent history of inference mismatch can serve as an endogenous signal for regulating the degree to which subsequent inference is constrained by learned internal dynamics during physical human--robot interaction. The high- and low-$w$ regimes showed the expected reconstruction--KL trade-off implied by the PV-RNN objective: weaker meta-prior constraint permitted greater posterior departure from the prior and lower reconstruction error. The more important finding is that this retrospective mismatch signal drove online regime transitions and that the resulting regulation was associated with consequences outside the reconstruction objective, including prospective prediction and robot-side physical interaction across different interaction structures.

\subsection{Reconstruction Error as a Higher-Order Regulatory Signal}
\label{sec:5:regulatory_signal}

The central contribution of the proposed framework is not the reduction of reconstruction error under low $w$ itself. From the variational objective in Eq.~\eqref{eq:loss}, reducing the meta-prior weakens the penalty on posterior--prior divergence and therefore gives posterior inference greater freedom to accommodate sensory observations. The observed combination of lower reconstruction error and larger raw KL divergence under low $w$ consequently verifies that the two predefined regimes produce the intended difference in posterior constraint.

The less direct result is that reconstruction mismatch operates at a second computational level. At the inference level, sensory mismatch drives posterior optimization within the current error-regression window. At the regulatory level, the temporal history of this mismatch determines the regime under which \emph{subsequent} inference is performed. Thus, the system uses the outcome of its own recent inference process as an endogenous signal for regulating future prior constraint.

This distinction is supported by effects that are not explicit terms of the reconstruction objective. Regime-dependent changes were observed in prospective prediction and robot-side physical interaction across all three tasks. Moreover, Task~3 showed that recent observations may be successfully accommodated while subsequent human motion still departs from the model's prior-generated trajectory. Thus, retrospective reconstruction mismatch can regulate subsequent inference without being equivalent to prospective prediction or physical coordination.

\subsection{Regulation Between Internally Generated and Externally Coupled Processing}
\label{sec:5:internal_external}

The results further indicate that internally generated and externally coupled processing should not be interpreted as competing states for which one regime is universally preferable. Rather, they provide different computational operating conditions. Stronger prior constraint preserves the influence of learned temporal dynamics, whereas weaker prior constraint permits greater accommodation of unexpected sensory input. The present study examines how selection between these operating conditions can be regulated online from the recent history of reconstruction mismatch.

The functional consequences of stronger sensory coupling also depend on
the reliability of sensory information \cite{Feldman2009, oyama2020integrated, ACC2022}. In a robotic system, sensor noise, malfunction, or communication loss could make direct sensory accommodation undesirable; under such conditions, higher $w$ could preserve reliable learned dynamics until trustworthy sensory information is restored. Conversely, persistent mismatch with reliable sensory observations may favor lower $w$ and stronger sensory accommodation. This interpretation is consistent with predictive processing and free energy accounts of the balance between prior expectations and sensory evidence \cite{Fri10b}.

PV-RNN provides an explicit computational handle on this balance through
the meta-prior \cite{Reza2019}. Previous embodied studies examined
behavior under fixed meta-prior settings \cite{hiroki2025,wirkuttis2023turn}; the present framework extends this principle by using endogenous reconstruction dynamics to select the processing regime online during interaction. Although the current implementation uses two discrete values, $w_L$ and
$w_H$, these regimes can be viewed as two operating regions along a
broader continuum between prior-dominated and sensory-responsive
inference.

\subsection{From Structured to Less-Constrained Human--Robot Interaction}
\label{sec:5:interaction_structure}

The progression from Tasks~1 and 2 to Task~3 examines meta-prior regulation across interaction conditions with different temporal organization. In the
structured tasks, lowering $w$ substantially reduced reconstruction error, prospective prediction error, and physical interaction conflict. Because the human-imposed trajectories were drawn from learned movement primitives, stronger sensory accommodation also supported accurate prediction of their subsequent evolution.

Task~3 showed that the same regulatory mechanism remained operative
without a prescribed movement sequence or transition schedule. Low $w$
again reduced reconstruction and prospective prediction errors while
increasing posterior--prior divergence, demonstrating that the mechanism
does not depend on the explicit PFSM structure used in Tasks~1 and 2.

However, Task~3 also revealed a partial separation between sensory accommodation and prospective coordination. Recent sensory observations could be successfully accommodated through posterior inference while subsequent human motion still departed from the model's prior-generated future trajectory. Reconstruction is retrospective and can improve after posterior adaptation to recent observations, whereas prospective coordination depends on anticipating future human motion. Future work could integrate the present inference-regulation mechanism with online adaptation of the generative model.

\subsection{Relation to Focused and Mind-Wandering Processing}
\label{sec:5:mindwandering}

The observed regulation also bears a functional relationship to distinctions between focused and mind-wandering processing. Mind wandering is associated with relative decoupling from immediate external sensory demands and increased internally generated content \cite{smallwood2015science,henriquez2016fluctuating,zukosky2021spontaneous}. Free-energy-based accounts have likewise examined how changes in the weighting of sensory evidence and internal beliefs may contribute to attentional transitions \cite{idei2024awareness}.

A previous PV-RNN study modeled autonomous transitions between externally
oriented and internally generated processing in simulated cognitive tasks
\cite{frontiers_oyama2025}. The present study extends this computational
principle to embodied physical interaction, where changes in the
prior--sensory balance have measurable consequences for latent dynamics,
reconstruction, prospective prediction, and physical interaction.

This correspondence is functional rather than phenomenological. The
high-$w$ regime is not equivalent to mind wandering, nor is low-$w$
processing equivalent to focused attention; the shared principle is the
regulation of the relative influence of internally generated dynamics and
external sensory information.

More broadly, future work could examine whether extending this continuum toward stronger internally generated processing provides a computational model for studying reduced sensory coupling, including functional analogues of imaginative or dream-like processing. Such extensions would remain computational hypotheses rather than claims concerning phenomenal consciousness or awareness. Such terminology, if used, refers only to functional differences in sensory coupling, consistent with recent cautions concerning consciousness claims in artificial systems \cite{Seth_2025}.

\subsection{Limitations and Future Directions}
\label{sec:5:limitations}

Several limitations define the scope of the present findings. First, although regime selection occurs online from endogenous reconstruction dynamics, the regulatory policy itself is manually specified rather than learned. The mechanism switches between two predefined meta-prior values using predefined reconstruction-error thresholds and a temperature parameter. Moreover, the present experiments were designed to characterize online regime selection rather than to establish its performance advantage over fixed-meta-prior operation. Direct comparisons with fixed-$w$ baselines will be necessary to determine whether dynamic regulation provides functional benefits over maintaining a single operating regime.

Second, error regression updates posterior inference but not the learned generative model parameters. Online adaptation of generative models has
been investigated in related predictive-coding frameworks \cite{sawada2026free}. A natural extension of the present work is therefore to integrate such model adaptation with the proposed inference-regulation mechanism, such that reconstruction dynamics regulate not only \emph{how} inference is performed but also \emph{when} model updating is engaged.

Third, reconstruction error does not explicitly represent sensory
reliability. A noisy or malfunctioning sensor could produce persistent
mismatch even though stronger sensory coupling would be undesirable.
Combining reconstruction dynamics with an estimate of sensory precision
could help distinguish unexpected but reliable observations from
unreliable sensory input.

Finally, the experiments used four arm-joint dimensions on a single humanoid platform and only four independently trained model realizations. Task~3 remained physically and kinematically constrained despite being less temporally structured than Tasks~1 and 2. Future evaluation should include additional human operators, independently collected interaction trajectories, richer sensory modalities, and higher-dimensional behavior. The relationship to internally generated cognitive processing discussed above therefore remains a functional computational interpretation that requires evaluation in broader behavioral settings.


\section{Conclusion}
\label{sec:conclusion}

This study introduced a predictive-coding-based mechanism for online regulation between internally generated and externally coupled processing during physical human--robot interaction. In a predictive-coding-inspired variational recurrent neural network, reconstruction error accumulated over recent interaction history was used to modulate the meta-prior and thereby regulate the relative influence of learned prior dynamics and
ongoing sensory information.

Across fixed structured, changing structured, and less-constrained interaction, reconstruction history drove online transitions between processing regimes with different degrees of prior constraint. The resulting lower reconstruction error and greater posterior--prior divergence under low $w$ are consistent with the PV-RNN objective; critically, regime selection was also associated with changes in prospective prediction and robot-side physical interaction, quantities not directly optimized by the reconstruction term.

Overall, the results show that temporally accumulated reconstruction mismatch can function not only as an inference signal but also as a higher-order signal that regulates the conditions of subsequent inference. Thus, the contribution lies in closing a feedback loop from recent inference mismatch to online regulation of prior constraint and demonstrating its consequences for prospective and embodied interaction dynamics.


\section*{Acknowledgments}
H. Oyama was supported by the Japan Society for the Promotion of Science (JSPS) KAKENHI Early-Career Scientists Grant No. 25K21307 and Grant-in-Aid for Transformative Research Areas (A) (Publicly Offered Research) Grant No. 26H00534. J. Tani was supported by the Japan Society for the Promotion of Science (JSPS) KAKENHI Grant-in-Aid for Transformative Research Areas (A), “Unified Theory of Prediction and Action,” Grant No. 26H01186.

\bibliographystyle{unsrt}  
\bibliography{library }

\end{document}